\documentclass[11pt,a4paper]{article}
\usepackage[hyperref]{emnlp2020}
\usepackage{times}
\usepackage{latexsym}

\usepackage{microtype}

\usepackage{graphicx}
\usepackage{booktabs,multirow}
\usepackage{caption}
\usepackage[whole]{bxcjkjatype}
\usepackage{bbding}
\usepackage{pifont}
\usepackage{pifont}

\usepackage{amssymb,mathtools}

\usepackage{bm}
\usepackage[normalem]{ulem} %
\def\dout{\bgroup
 \markoverwith{\lower-0.2ex\hbox
 {\kern-.03em\vbox{\hrule width.2em\kern0.45ex\hrule}\kern-.03em}}%
 \ULon}
\MakeRobust\dout
\usepackage{url}
\usepackage{xspace}
\usepackage{lingmacros}
\usepackage{tabularx}
\usepackage{longtable}
\usepackage{scrextend}
\usepackage{algorithm}
\usepackage{algorithmic}
\usepackage{subcaption}

\aclfinalcopy %
\def\aclpaperid{131} %

\newcommand{\argmax}{\mathop {\rm argmax}\limits}

\title{\textit{Langsmith}: An Interactive Academic Text Revision System}

\author{Takumi Ito\thanks{$\quad$The authors contributed equally}$^{\;\;,1,2}$ , Tatsuki Kuribayashi\footnotemark[1]$^{\;\;,1,2}$,
Masatoshi Hidaka\footnotemark[1]$^{\;\;,3}$, \\ {\bf Jun Suzuki$^{1,4}$, and Kentaro Inui$^{1,4}$}  \\
 $^1$Tohoku University 
 $^2$Langsmith Inc.
 $^3$Edge Intelligence Systems Inc.
 $^4$RIKEN \\
 \texttt{\{t-ito, kuribayashi, jun.suzuki, inui\}@ecei.tohoku.ac.jp}\\ \texttt{hidaka@edgeintelligence.jp} \\}

\begin{document}
\maketitle
\begin{abstract}
Despite the current diversity and inclusion initiatives in the academic community, researchers with a non-native command of English still face significant obstacles when writing papers in English.
This paper presents the \textit{Langsmith} editor, which assists inexperienced, non-native researchers to write English papers, especially in the natural language processing (NLP) field.
Our system can suggest fluent, academic-style sentences to writers based on their rough, incomplete phrases or sentences.
The system also encourages interaction between human writers and the computerized revision system.
The experimental results demonstrated that Langsmith helps non-native English-speaker students write papers in English.
The system is available at \url{https://emnlp-demo.editor.langsmith.co.jp/}.
\end{abstract}

\section{Introduction}
\label{sec:intro}

Currently, diversity and inclusion in the natural language processing (NLP) community are encouraged.
In fact, at the latest NLP conference at the time of writing\footnote{The 58th Annual Meeting of the Association for
Computational Linguistics}, papers were submitted from more than 50 countries.
However, one obstacle can limit this diversity: \textit{The papers must be written in English.}
Writing papers in English can be a daunting task, especially for inexperienced, non-native speakers.
These writers often struggle to put their ideas into words.

To address this problem, we built the \textit{Langsmith} editor, an assistance system for writing NLP papers in English.\footnote{See \url{https://www.youtube.com/channel/UCjHeZPe0tT6bWxVVvum1bFQ} for the screencast.}
The main feature in Langsmith is a revision function, which suggests fluent, academic-style sentences based on writers’ rough, incomplete drafts.

The drafts might be so rough that it becomes challenging to understand the user's intended meaning to use as inputs. 
In addition, several potentially plausible revisions can exist for the drafts, especially when the input draft is incomplete.

\begin{figure}[t]
    \includegraphics[width=\hsize]{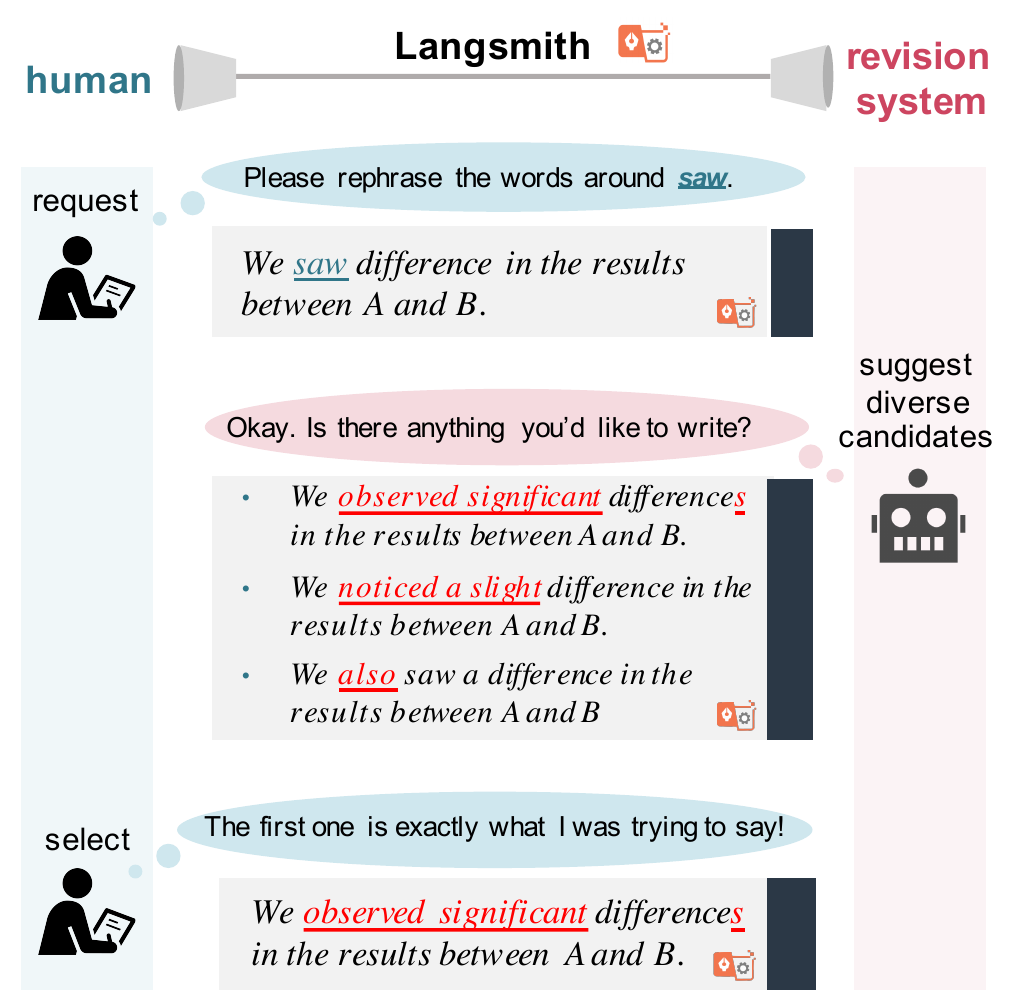}
          \centering
\vskip -2mm
    \caption{An overview of interactively writing texts with a revision system.}
    \label{fig:intro}
\end{figure}

Based on such difficulties, our system provides two ways for users to customize the revision: the users can (i) request specific revisions, and (ii) select a suitable revision from diverse candidates (Figure~\ref{fig:intro}).
In particular, the request stage allows users to specify the parts that require intensive revision.

\begin{figure*}[t]
    \includegraphics[width=15cm]{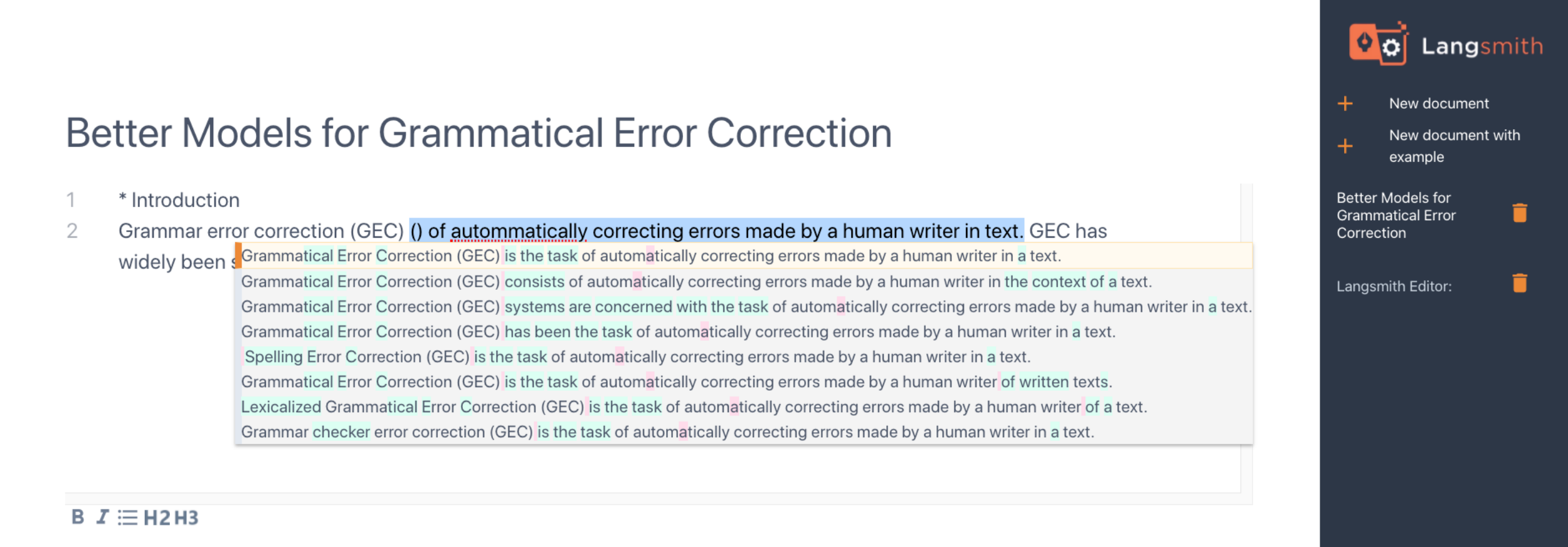}
          \centering
    \caption{Screenshot of Langsmith. The revision feature suggests various revisions for the input ``\textit{Grammar error correction (GEC) \texttt{()} of automatically correcting errors made by a human writer in text.}'' The characters highlighted in green are added to the original sentence, and the red points indicate tracked deletions.}
    \label{fig:demo_overview}
\end{figure*}

Our experiments demonstrate the effectiveness of our system.
Specifically, students whose first language is Japanese, which differs greatly from English, managed to write better drafts when working with Langsmith.

Langsmith has other assistance features as well, such as text completion with a neural language model.
Furthermore, the communication between the server and the web frontend is achieved via a protocol specialized in writing software called the Text Editing Assistance Smartness Protocol for Natural Language (TEASPN)~\cite{hagiwara-etal-2019-teaspn}.
We hope that our system will help the NLP community and researchers, especially those lacking a native command of English.\footnote{This paper was also written using Langsmith.}

\section{Related work}
\label{sec:rel}

\subsection{Natural language processing for academic writing}
\label{subsec:rel:academic}

Academic writing assistance has gained considerable attention in NLP~\cite{wu-etal-2010-automatic-collocation,yimam2020automatic, lee-webster-2012-corpus}, and several shared tasks have been organized~\cite{dale2011helping,daudaravicius2015aesw}.
These tasks focus on polishing texts in already published articles or documents near completion.
In contrast, this study focuses on revising texts in the earlier stages of writing (e.g., first drafts), where inexperienced, non-native authors might even struggle to convey their ideas accurately.

\citet{Ito_2019} introduced a dataset and models for revising early-stage drafts, and the 1-to-N nature of the revisions was pointed out. 
We tackled this difficulty by designing an overall demonstration system, including a user interface.

\subsection{Writing assistance tools}
\label{subsec:rel:tool}

\paragraph{Error checkers.}
Grammar/spelling checkers are typical writing assistance tools.
Some highlight errors (e.g., \textit{Write\&Improve}\footnote{\url{writeandimprove.com}}), while others suggest corrections (e.g., \textit{Grammarly\footnote{\url{https://www.grammarly.com}}}, \textit{LanguageTool}\footnote{\url{https://languagetool.org}}, \textit{Ginger\footnote{\url{https://www.gingersoftware.com}}}, and \textit{LinggleWrite}; see~\citet{tsai-etal-2020-lingglewrite}) for writers.

Langsmith has a revision feature~\cite{Ito_2019}, as well as a grammar/spelling checker.
The revision feature suggests better versions of poor written phrases or sentences in terms of fluency and style, whereas error checkers are typically designed to correct apparent errors only.
In addition, Langsmith is specialized for the NLP domain and enables domain-specific revisions, such as correcting technical terms.

\paragraph{Text completion.}
Completing a text is another typical feature in writing assistance applications  (\textit{WriteAhead}\footnote{\url{writeahead.nlpweb.org}}, \textit{Write With Transformer}\footnote{\url{https://transformer.huggingface.co}}, and \textit{Smart Compose}; see~\citet{Chen:2019}).
Our system also has a completion feature, which is specialized in academic writing (e.g., completing a text based on a section name).

\section{The \textit{Langsmith} editor}
\label{sec:langsmith}

\subsection{Overview}
\label{subsec:langsmith:overview}

This section presents Langsmith, a web-based text editor for academic writing assistance (Figure~\ref{fig:demo_overview}).
The system has the following three features: (i) text revision, (ii) text completion, and (iii) a grammatical/spelling error checker.
These features are activated when users select a text span, type a word, or push a special key.

As a case study, this work focuses on paper writing in the NLP domain.
Thus, each assistance feature is specialized in the NLP domain.
The following sections explain the details of each feature.

\subsection{Revision feature}
\label{subsec:langsmith:revision}

The revision feature, the main feature of Langsmith, suggests better sentences in terms of fluency and style for a given draft sentence (Figure~\ref{fig:demo_overview}).
This feature is activated when the user selects a sentence or smaller unit.

Writers sometimes struggle to put their ideas into words.
Thus, the input draft for the revision systems can be incomplete, or less informative.
Based on such a challenging situation, we examine the \textsc{Request} and \textsc{Select} framework to help users discover sentences that better match what the user wanted to write.

\begin{figure}[!t]
\centering
\subfloat[Revisions focusing on \textit{This formulation $\cdots$ and output.}]{\includegraphics[clip, width=3.0in]{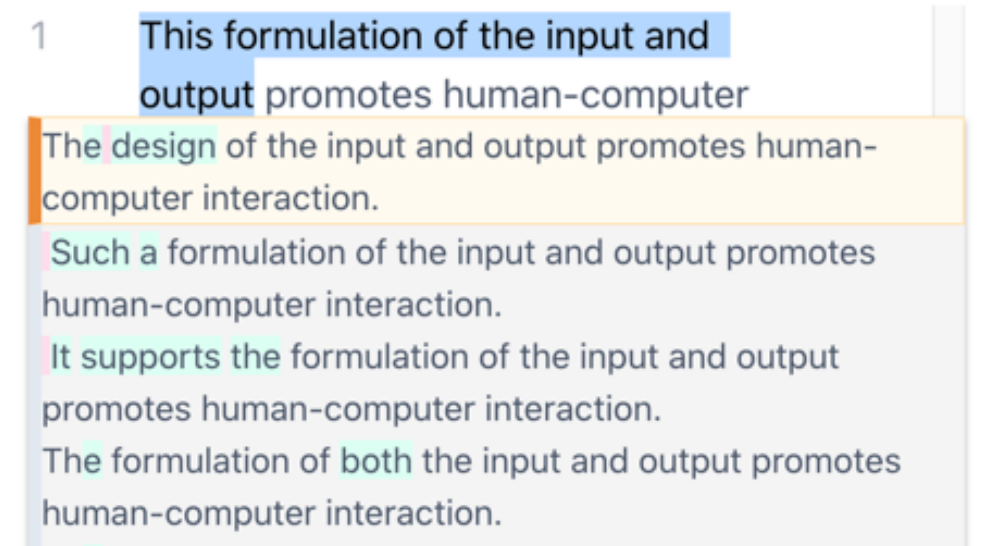}
\label{fig:label-A}}
\\
\subfloat[Revisions focusing on \textit{promote.}]{\includegraphics[clip, width=3.0in]{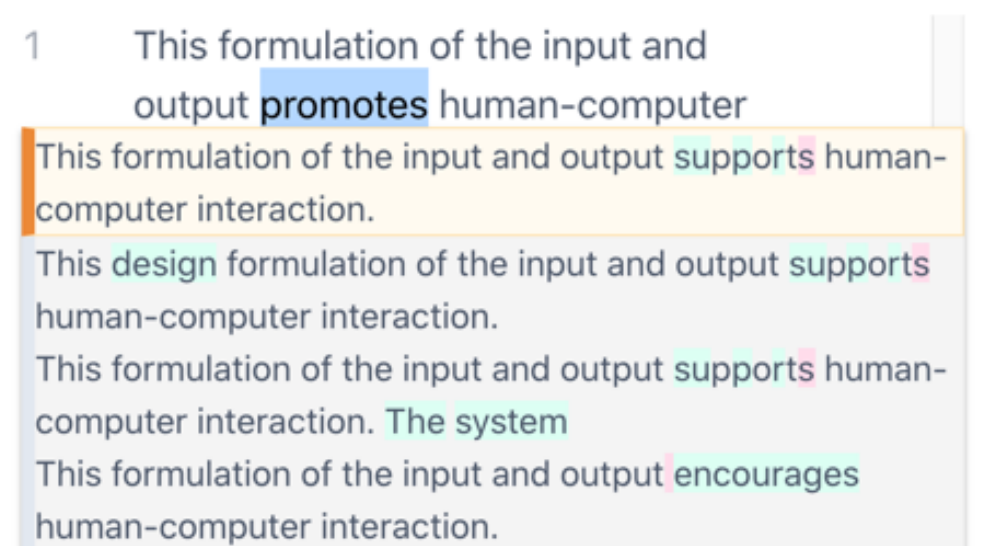}
\label{fig:label-B}}
\\
\subfloat[Revisions focusing on \textit{human--computer interaction.}]{\includegraphics[clip, width=3.0in]{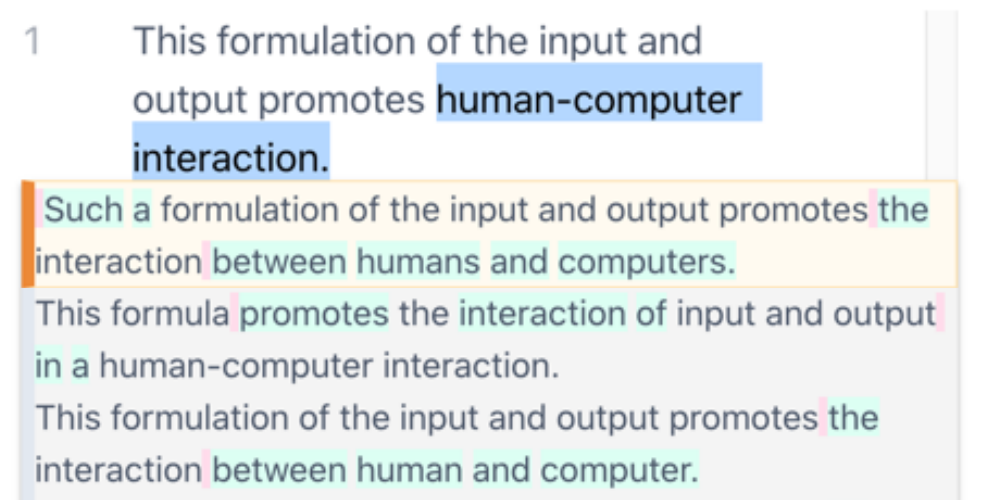}
\label{fig:label-C}}
\caption{The focus of the revision depends on the parts selected by users.}
\label{fig:edit_points}
\end{figure}

\paragraph{\textsc{Request} stage.} 
Langsmith provides two ways for users to request a specific revision, which can prevent unnecessary revisions being provided to the user. 

First, users can specify where the system should intensively revise a text.\footnote{The system performs sentence-level revisions. Hence the users are instructed to select the non-sentence-crossing area.}
That is, when a part of a sentence is selected, the system intensively rephrases the words around the selected part.\footnote{We allow the system to correct the parts outside the selected span because sometimes the revision for a specific part requires another adjustment for the other parts.
}
Figure~\ref{fig:edit_points} demonstrates the change of the revision focus, depending on the selected text span.
Note that controlling the revision focus was not explored in the original sentence-level revision task~\cite{Ito_2019}.
This feature is also inspired by~\citet{grangier-auli-2018-quickedit}.

Second, users can insert placeholder symbols, ``\texttt{()}'', at specific points in a sentence.
The system revises the sentence by replacing the symbol with an appropriate expression regarding its context.
The input for the revision in Figure~\ref{fig:demo_overview} also has the placeholder symbol.
Here, for example, the symbol is replaced with ``the task.''
This feature is inspired by~\citet{wanrong2019textinfilling,donahue-etal-2020-enabling,Ito_2019}.

\paragraph{\textsc{Select} stage.}
The system provides several revisions (Figure~\ref{fig:demo_overview}).
Note that there is typically more than one plausible revision in terms of fluency and style, in contrast to correcting surface-level errors~\cite{napoles2017jfleg}.

The diversity of the output revisions is encouraged using diverse beam search~\cite{journals/corr/VijayakumarCSSL16}.
In addition, these revisions are ordered by a language model that is fine-tuned for NLP papers.
That is, revisions with lower perplexity are listed in the upper part of the suggestion box.
Furthermore, the revisions are highlighted in colors, which makes it easier to distinguish the characteristics of each revision.

\paragraph{Implementation.}
We trained a revision model using LightConv~\cite{wu2018pay} implemented in Fairseq~\cite{ott2019fairseq}.
The revision model generates a sentence based on a given input sentence.
The model was trained on a slightly modified version of the synthetic training data used in~\citet{Ito_2019}.
As an example of these modifications, synthetic edit marks were added for a subset of the training data. 
These marks were attached to a part of the input sentence that has many edits compared to its reference.\footnote{Special symbols are attached at the beginning and the end of the specific subsequence.}
Thus, the marks can provide a hint for the system to determine where to edit.
When using Langsmith, the marks are attached to the span selected by the users.
The system is expected to intensively revise the wording in the specified span.
Details are in Appendix~\ref{app:revision}.

\begin{figure}[t]
    \includegraphics[width=7.5cm]{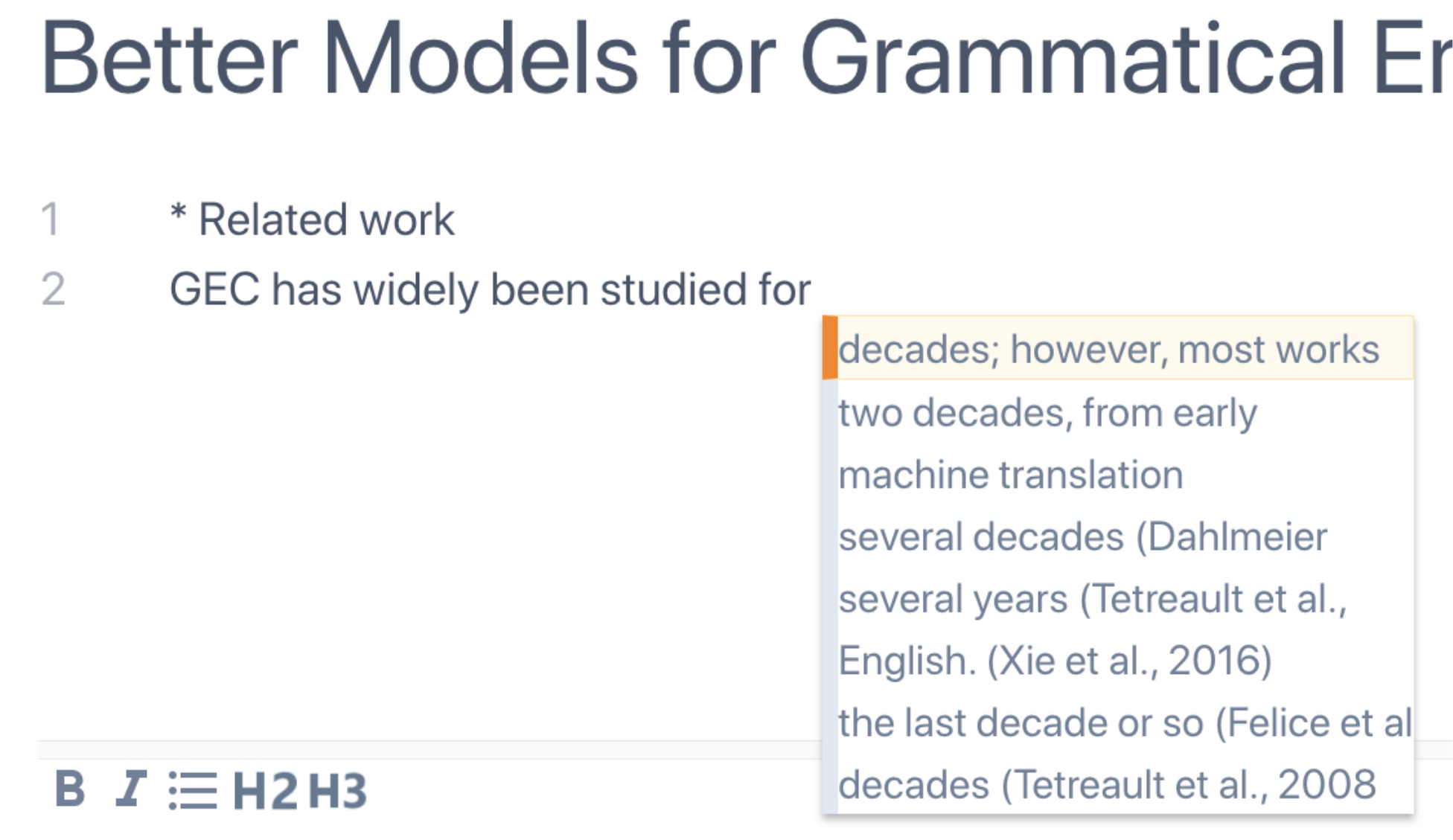}
          \centering
    \caption{An example of the completion feature. These suggestions are conditioned by the left context, section name (\textit{Related work}) and the paper title (\textit{Better Models for Grammatical Error Correction.})}
    \label{fig:completion}
\end{figure}

\begin{figure}[t]
    \includegraphics[width=7.5cm]{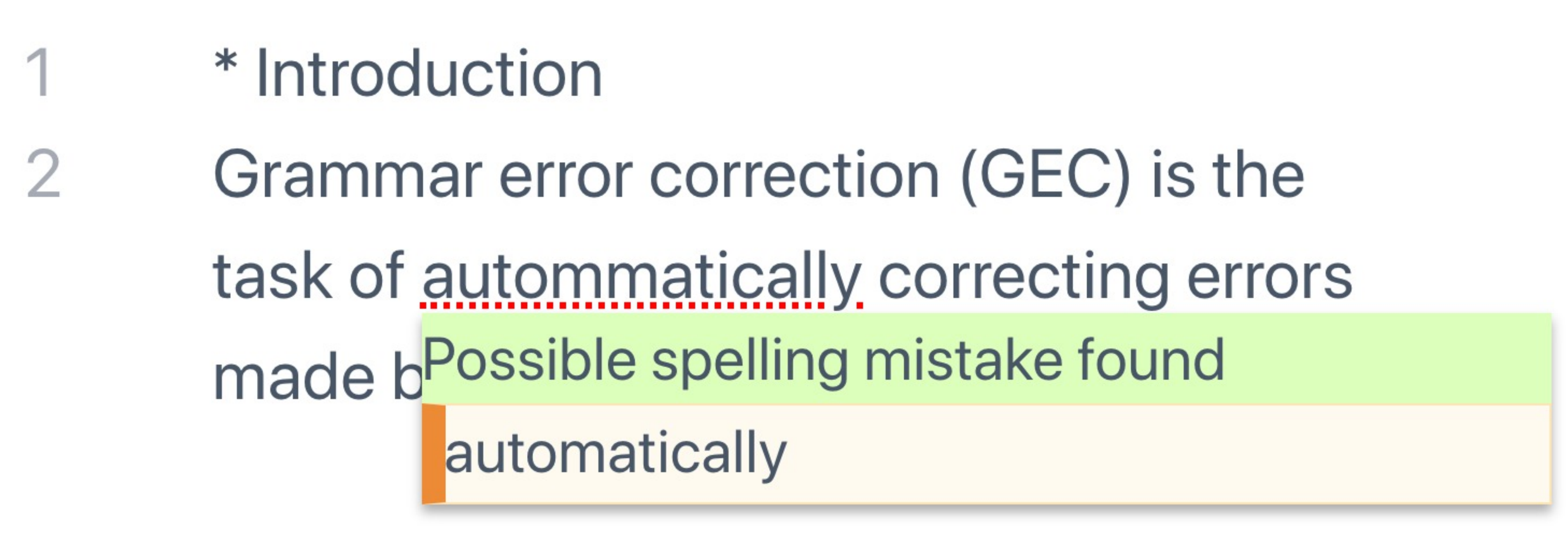}
          \centering
    \caption{The interface of the error correction feature. Errors are automatically highlighted with a red line. The corrections are suggested when the user hovers over the highlighted words.}
    \label{fig:error}
\end{figure}

\subsection{Other features}
\label{subsec:langsmith:other}

\paragraph{Completion feature.} 
When the user presses the \texttt{Tab} key, the completion feature generates plausible preceding phrases from the cursor point (Figure~\ref{fig:completion}).
This feature can consider the paper title and section name as well as the text to the left of the cursor.

We used GPT-2 small (117M)~\cite{radford2019language} fine-tuned on the papers collected from the ACL Anthology\footnote{\url{https://www.aclweb.org/anthology}}.
Paper titles and section names were concatenated at the beginning of the corresponding paragraphs in the fine-tuning data.
Details are in Appendix~\ref{app:completion}.

\paragraph{Error correction feature.}
We used LanguageTool,\footnote{\url{https://github.com/languagetool-org/languagetool/releases/tag/v3.2}} an open-source grammatical/spelling error correction tool. 
Each time the text changes, this feature is called upon.
The detected errors are then automatically highlighted with red lines (Figure~\ref{fig:error}).%
The corrections are listed when the user hovers over the highlighted words.

\subsection{Protocol}
\label{subsec:langsmith:protocol}
Langsmith was developed based on the TEASPN Software Development Kit~\cite{hagiwara-etal-2019-teaspn}.\footnote{\url{https://github.com/teaspn/teaspn-sdk}}
TEASPN defines a set of APIs for writing software (e.g., text editors) to communicate with servers that implement NLP technologies (e.g., revision model).
We extended the protocol to convey title and section information in the completion feature.
Since Langsmith is a browser-based tool and frequently communicates with a web server running models, we used WebSocket to achieve smooth communication.

\section{Experiments and results}
\label{sec:exp}

We demonstrate the effectiveness of human--machine interactions in revising drafts implemented in our system.
We also check whether the \textsc{Request} stage in the revision feature works adequately.

\subsection{On the revised draft quality}
\label{subsec:exp:quality}

\paragraph{Settings.}
We suppose a situation where a person writes a draft in their native language (non-English language), translates it to English, and then revises it further to create an English-language draft. %
In order to simulate this situation, we first collected Japanese-language version of the abstract sections from eight Japanese peer-reviewed journals.\footnote{We used the journals accepted at \url{https://www.anlp.jp/en/index.html}.}
Then, the abstracts were translated into English with an off-the-shelf translation system\footnote{\url{https://translate.google.co.jp}}.
We considered the translated abstracts as first drafts.
The task is to revise the first drafts.
Expert translators created reference final drafts from the Japanese versions of the drafts.\footnote{We used~\url{https://www.ulatus.com/}.}
We evaluated the quality of the revised versions by comparing them with the corresponding final drafts.

We compared three versions of revised drafts to evaluate the effectiveness of Langsmith:

\vspace{-0.25cm}
\begin{itemize}
    \setlength{\parskip}{0cm}
    \setlength{\itemsep}{0cm}
    \item one fully and automatically revised by Langsmith (\textsc{Machine-only} revision)
    \item one revised by a human writer without Langsmith (\textsc{Human-only} revision), and
    \item one revised by a human writer using assistance features in Langsmith (\textsc{Human\&Machine} revision).
\end{itemize}
\vspace{-0.25cm}

\noindent
The following paragraphs explain how we obtained the above three versions of the revisions.
Appendix~\ref{app:stats} shows the statistics of the drafts.

\begin{table}[t]
    \centering
    \renewcommand{\arraystretch}{0.9}
    \begin{tabular}{lc}
    \toprule
    Condition & BLEURT \\
    \cmidrule(r){1-1} \cmidrule(l){2-2}
    \textsc{Human\&Machine}  & \textbf{-0.45} \\
    \textsc{Human-only}  & -0.51 \\
    \textsc{Machine-only} & -0.51 \\ 
    \cmidrule(r){1-1} \cmidrule(l){2-2}
    First drafts & -0.70 \\
    \bottomrule
    \end{tabular}
 \caption{Comparison of the revision quality. The scores are averaged over the corresponding revisions. Higher scores indicate that the drafts are closer to the final drafts.}
 \label{tbl:results}
\end{table}

\paragraph{\textsc{Machine-only} revision.}
We automatically applied the revision feature to the drafts (each sentence) without the \textsc{Request} and \text{Select} stages.
For each sentence, the revision with the highest generation probability was selected.\footnote{The hyperparameters for decoding revisions were the same as the revision feature in Langsmith. Re-ranking with the language model was also employed.}
We created one \textsc{Machine-only} revision for each first draft.

\paragraph{\textsc{Human-only} revision.}
Human writers revise a given first draft.
The writers can only access to the error correction feature.
This setting simulates the situations that writers typically face.

\paragraph{\textsc{Human\&Machine} revision.}
Human writers revise a given first draft with full access to the Langsmith features.

\paragraph{Human writers.}
We asked 16 undergraduate and master's students at an NLP laboratory to revise the first drafts in terms of fluency and style.
The students were Japanese natives, representatives of the inexperienced researchers in a country where the spoken language is considerably different from English.
Each participant revised two different first drafts, one with the \textsc{Human-only} setting and the other one with the \textsc{Human\&Machine} setting.

Half of the participants first revised a draft with the \textsc{Human-only} setting, and then revised another draft with the \textsc{Human\&Machine} setting; the other half performed the same task in the opposite order.
Ultimately, we collected two \textsc{Human\&Machine} revisions and two \textsc{Human-Only} revisions for each first draft.

\paragraph{Comparison and results.}
We compared the quality of the three versions of the revised drafts: \textsc{Machine-Only} revision, \textsc{Human-Only} revision, and  \textsc{Human\&Machine} revision.
We compared the revised drafts with their corresponding final draft using BLEURT~\cite{sellam-etal-2020-bleurt}, the state-of-the-art automatic evaluation metric for natural language generation tasks.
Details of the evaluation procedure is shown in Appendix~\ref{app:eval}.
Note that the score is not in the range $[0,1]$, and a higher score means that the revision is closer to the final draft.
Table~\ref{tbl:results} shows that \textsc{Human\&Machine} revisions were significantly better\footnote{We applied a bootstrap hypothesis test~\cite{W04-3250}, and the score of \textsc{Human\&Machine} was significantly higher than the \textsc{Human-only} and \textsc{Machine-only} scores ($p < 0.05$).} than \textsc{Machine-only} and \textsc{Human-only} revisions.
The results suggest the effectiveness of human--machine interaction achieved in Langsmith.
Since this experiment was relatively small in scale and only used an automatic evaluation metric, we will conduct a larger-scale experiment with human evaluations in the future.

\subsection{User study}

\begin{table}[t]
    \centering
    \renewcommand{\arraystretch}{0.9}
    \begin{tabular}{cp{1cm}p{1cm}p{1.4cm}p{1.4cm}}
    \toprule
     Q. & Strongly agree & Slightly agree & Slightly disagree & Strongly disagree \\
    \cmidrule(r){1-1} \cmidrule(l){2-2} \cmidrule(l){3-3} \cmidrule(l){4-4} \cmidrule(l){5-5}
     (I) & 87.5 & 12.5 & 0 & 0 \\
     (II) & 50.0 & 50.0 & 0 & 0 \\
     (III) & 62.5 & 31.3 & 6.3 & 0 \\
     (IV) & 12.5 & 50.0 & 31.3 & 6.3 \\
     (V) & 75.0 & 12.5 & 6.3 & 6.3 \\
     (VI) & 43.8 & 43.8 & 12.5 & 0 \\
    \bottomrule
    \end{tabular}
 \caption{Results of the user study about (I)-(VI). The scores denote the percentage of the participants who chose the option.}
 \label{tbl:answer_q1_6}
\end{table}

\begin{table}[t]
    \centering
    \begin{tabular}{cc}
    \toprule
    Feature & percentage\\
    \cmidrule(r){1-1} \cmidrule(l){2-2}
     revision & 100 \\
     completion & 31.3 \\
     correction & 62.5 \\
    \bottomrule
    \end{tabular}
 \caption{Results of the user study about helpful features. The scores denote the percentage of the participants who chose the feature (multiple choice question).}
 \label{tbl:answer_q7}
\end{table}

After the experiments outlined in Section~\ref{subsec:exp:quality}, we asked the participants about the usability of Langsmith.
The 16 participants were instructed to evaluate the following statements:

\begin{itemize}
    \setlength{\parskip}{0cm}
    \setlength{\itemsep}{0cm}
    \item[(I)$\,\,$] Langsmith was more helpful than the Baseline environment for the revision task.
    \item[(II)$\,$] Comparing the text written by the two environments, the text written with Langsmith was better.
    \item[(III)] The feature of specifying where to intensively revise was helpful.
    \item[(IV)] The placeholder feature in the revision feature was helpful.
    \item[(V)$\,$] Providing more than one output from the revision feature was helpful.
    \item[(VI)] Providing more than one output from the completion feature was helpful.
\end{itemize}

\noindent
The participants evaluated the statements (I)-(VI) on a four-point scale: (a) strongly agree, (b) slightly agree, (c) slightly disagree, and (d) strongly disagree. 
In addition, the participants answered whether each feature was helpful in writing.

\paragraph{Results.}

Tables~\ref{tbl:answer_q1_6} and~\ref{tbl:answer_q7} show the results of our user study.
From the responses to (I) and (II), we observed that the users were satisfied with the writing experience with Langsmith.
The responses to (III), (IV), and (V) support the idea that our \textsc{Request} and \textsc{Select} stages are helpful.
Here, using the place holders was relatively not helpful.
The responses to (VI) also suggest that showing several candidates does not bother the users. 
Table~\ref{tbl:answer_q7} displays the result of whether each feature was helpful in writing.
The result indicates that the revision feature was the most useful for creating drafts using the implemented features.

\subsection{Sanity check of the \textsc{Request} stage}

Finally, we checked the validity of our method to control the revision based on the selected part of the sentence (Figure~\ref{fig:edit_points}).

\paragraph{Settings.}
We randomly collected 1,000 sentences from the first drafts created with the translation system.
In each sentence with $T$ tokens $x=(w_1,\cdots,w_T)$, we randomly inserted edit marks to specify a certain span $s=(i,j)$ in $x$ $(1 \leq i < j \leq T, 1 \leq j-i \leq 5)$.
Specifically, special tokens were inserted before $w_i$ and after $w_j$ in $x$.
We denote the input sentence with these edit marks as $x^{\text{edit}}$.
We then obtained 10-best outputs of the revision system $(y^{\text{edit}}_1, \cdots, y^{\text{edit}}_{10})$ for each $x^{\text{edit}}$.
Here, these output sentences were generated through the diverse beam search with the same settings as the revision feature in Langsmith.
We calculated the following score for each input sentence and its revisions:

\vspace{-0.6cm}
\begin{align}
\label{eq:edit_change}
\nonumber r = |\{y^{\text{edit}}_k \;|\; x_{i:j} \in \text{ngram}(y^{\text{edit}}_k), \nonumber 1\leq k \leq 10\}|
\end{align}
\vspace{-0.6cm}

\noindent
where $x_{i:j}$ denotes the subsequence $(w_i,\cdots,w_j)$ in $x$.
The function $\text{ngram}(\cdot)$ returns a set of all the n-grams of a given sequence.
A lower $r$ indicates that the subsequence specified with the edit marks are more frequently rephrased.

We also obtained a score $r^{\prime}$ for each $x$. 
$r^{\prime}$ was calculated using the input without the edit marks $x$ and its 10-best outputs $y_k$.
We compared $r$ and $r^{\prime}$ for each $x$.

\paragraph{Results.}
We observed that $r$ frequently\footnote{We conducted the one-side sign test. The difference is significant with $p \leq 0.05$.} had lower values than $r^{\prime}$.
That is, a certain subsequence was more rephrased by the revision system when it had the edit marks than when it did not.
These results validate our approach of controlling the revision focus, which is implemented in the \textsc{Request} stage of the revision feature.

\section{Conclusions}
We have presented Langsmith, an academic writing assistance system.
Langsmith provides a writing environment, in which human writers use several assistance features to improve the quality of texts.
Our experiments suggest that our system is useful for inexperienced, non-native writers in revising English-language papers.
We are aware that our experimental settings were not fully well-designed (e.g., we had only Japanese participants, and no human evaluation).
We will evaluate Langsmith in more sophisticated settings.
We hope that our system contributes to breaking language barriers in the academic community.

\section*{Acknowledgement}
We are grateful to Ana Brassard for her feedback on English.
We also appreciate the participants of our user studies.
This work was supported by Grant-in-Aid for JSPS Fellows Grant Number JP20J22697.

\bibliography{emnlp2020}
\bibliographystyle{acl_natbib}

\clearpage

\appendix
\section{Details on revision model}
\label{app:revision}

\paragraph{Data.}
We trained the revision model using the slightly modified version of the synthetic training data introduced in~\citet{Ito_2019}.
They created several types of synthetic training data with several noising methods; (i) heuristic noising method, (i) grammatical error generation, (iii) style removal, and (iv) entailed sentence generation.
We used the data created by the heuristic noising method, style removal, and the entailed sentence generation for training the revision model.
Note that we did not use the data generated by the grammatical error generation because grammatical error correction feature was implemented separately from the revision feature in Langsmith.

We attached the edit marks to the subpart of the training data generated by the style removal method.
Let $x_{1:N}=(x_1, x_2, \cdots, x_N)$ and $y_{1:T}=(y_1, y_1, \cdots, y_M)$ be an input sentence with $N$ tokens and its revision with $M$ tokens, respectively.
Here $x$ was the synthetic draft sentence generated by the style removal method from $y$.
The training dataset consists of the pairs of ($x$, $y$).

For each $(x,y)$, we first determined if each word in $x$ was rewritten compared to $y$.
We assumed that a token $x_i \in x$ was rewritten if a token with the same lemma as $x_i$ was not in $\{y_j|\text{max}(0,i-3) \leq j \leq \text{min}(M,i+3)\}$.
Here we obtained a sequence $c\in\{0,1\}^N$, where each element $c_i$ corresponds to whether the token $x_i$ was rewritten or not.
If $x_i$ was written in $y$, $c_i$ is 1; otherwise $c_i$ is 0.
Then, we defined a score $r(c)$ for each $(x,y)$ as follows:

\vspace{-0.2cm}
\begin{equation}
    \nonumber r(c) = \frac{\sum_{i=1}^N c_i}{|c|}
\end{equation}
\vspace{-0.2cm}

\noindent
where $|\cdot|$ returns the length of the vector.
If $r(c) > 0.4$, we did not attach the edit marks.

When $r(c) \leq 0.4$, we obtained a span $s=(a,b)$ for $x$ and $c$ as follows:
\begin{eqnarray}
\nonumber \argmax_{(a,b) \in \mathcal{S}}&& \hspace{-0.7cm} \sum_{i=a}^b c'_i - \sum_{i=0}^{a-1} c'_i - \sum_{i=b+1}^{N+1} c'_i \\
\nonumber \text{where} \;\; 
c'_i &=& \begin{cases}
    10 & (c_i=1) \\
    0 & (i=0,N+1) \\
    -1 & (\text{otherwise})
  \end{cases} \\
\nonumber \mathcal{S} &=& \{(a, b)\;|\;a,b \in {1,\cdots, N}, a \leq b\} 
\end{eqnarray}

\noindent
Based on the obtained $s=(a,b)$, we inserted \texttt{<?} before the token $x_a$, and \texttt{?>} after the token $x_b$.
We included the data with special symbols added by such a procedure in the training data.

When the users select a subsequence of a sentence in Langsmith, the edit marks are attached to the input sentence.
For example, if the user selects a span ``promote'' in the sentence ``This formulation of the input and output promotes human-computer interaction.'', the input to the revision feature is formatted as follows: \texttt{This formulation of the input and output <? promotes ?> human-computer interaction.}

\paragraph{Model.}
Table~\ref{tbl:hyperparam_revison} shows the hyperparameters of the revision model.
In the decoding phase, we used the diverse beam search~\cite{journals/corr/VijayakumarCSSL16}.
Beam size is set to 15.
The diverse beam group and the diverse beam strength are 15 and 1.0, respectively.

Specifically, we first obtained top-15 hypotheses, and then these hypotheses were re-ranked by the language model.
Here, the language model considers 20 tokens in the left context and 20 tokens in the right context beyond the sentence.
We excluded the hypotheses with a perplexity greater than 1.3 times the perplexity of the input.
We finally showed the top-8 revisions re-ranked to the users.
The language model used for re-ranking is the same as the model used for the completion feature (Appendix~\ref{app:completion}).

\begin{table*}[t]
    \centering
    \begin{tabular}{llc} \toprule
     \multirow{1}{*}{Fairseq model} & architecture & lightconv\_iwslt\_de\_en \\
    \cmidrule(lr){1-1} \cmidrule(lr){2-2} \cmidrule(lr){3-3}
    \multirow{5}{*}{Optimizer} & algorithm & Adam \\
    & learning rate & 5e-4 \\
    & adam epsilon & 1e-08 \\
    & adam betas & (0.9, 0.98) \\
    & weight decay & 0.0001 \\
    & clip norm & 0.0 \\
    \cmidrule(lr){1-1} \cmidrule(lr){2-2} \cmidrule(lr){3-3}
    \multirow{4}{*}{Learning rate scheduler} & type & inverse\_sqrt \\
    & warmup updates & 4000 \\
    & warmup init lrarning rate & 1e-7 \\
    & min learning rate & 1e-9 \\
    \cmidrule(lr){1-1} \cmidrule(lr){2-2} \cmidrule(lr){3-3}
    \multirow{2}{*}{Training} & batch size & 24,000 tokens \\ 
    & updates & 1,050,530 steps \\  \bottomrule
        \end{tabular}
        \caption{Hyperparameters of the revision feature.}
        \label{tbl:hyperparam_revison}
\end{table*}

\section{Details on completion model}
\label{app:completion}

\paragraph{Data.}
We collected 234,830 PDFs of the papers published in ACL Anthology by 2019.
We used GROBID\footnote{\url{https://github.com/kermitt2/grobid}} for extracting the text information from the PDF files.
The training data is formatted as shown in Table~\ref{tbl:data_completion}.
The title name is omitted with 20\% probability.
The order of the sections in the same paper was shuffled.

\begin{table}[t]
    \centering
    \begin{tabular}{l} \toprule
    @ Title @ \\
    \\
    * Section name \\
    Texts in the section \\
    $\cdots$ \\
    \\
    * Section name \\
    Texts in the section \\
    $\langle$\textbar endoftext\textbar $\rangle$ \\
    \\
    @ Title (of another paper) @ \\
    $\cdots$ \\
    \bottomrule
        \end{tabular}
        \caption{The format of the training data for the completion model.}
        \label{tbl:data_completion}
\end{table}

\paragraph{Model.}
We used a pre-trained GPT-2 small (117M).
Table~\ref{tbl:hyperparam_completion} shows the hyperparameters for fine-tuning the pre-trained GPT-2.
We used an implementation in Transformers~\cite{Wolf2019HuggingFacesTS}.
We used nucleus sampling~\cite{Holtzman2020The} with $p=0.97$ to generate the texts.

\begin{table*}[t]
    \centering
    \begin{tabular}{llc} \toprule
     \multirow{1}{*}{Model} & architecture & gpt2 \\
    \cmidrule(lr){1-1} \cmidrule(lr){2-2} \cmidrule(lr){3-3}
    \multirow{5}{*}{Optimizer} & algorithm & Adam \\
    & learning rate & 5e-5 \\
    & adam epsilon & 1e-8\\
    & adam betas & (0.9, 0.999) \\
    & weight decay & 0.0 \\
    & clip norm & 1.0 \\
    \cmidrule(lr){1-1} \cmidrule(lr){2-2} \cmidrule(lr){3-3}
    \multirow{4}{*}{Learning rate scheduler} & type & linear \\
    & warmup updates & 0 \\
    & max learning rate & 5e-5 \\
    & total epochs (just used for scheduling) & 100\\
    \cmidrule(lr){1-1} \cmidrule(lr){2-2} \cmidrule(lr){3-3}
    \multirow{2}{*}{Training} & batch size & 262,144 tokens \\ 
    & updates & 138,300 steps \\  \bottomrule
        \end{tabular}
        \caption{Hyperparameters for fine-tuning LMs.}
        \label{tbl:hyperparam_completion}
\end{table*}

\section{Statistics of the drafts}
\label{app:stats}

Table~\ref{tbl:stats} shows the statistics of the drafts collected in Section~\ref{sec:exp}.
The column ``word type'' shows the number of types of the tokens used in the drafts.

\begin{table*}[t]
    \centering
    \renewcommand{\arraystretch}{0.9}
    \begin{tabular}{lcc}
    \toprule
    drafts & length & word types \\
    \cmidrule(r){1-1} \cmidrule(l){2-2} \cmidrule(l){3-3}
    Final drafts (reference) & 199 $\pm$ 52 & 108 $\pm$ 17\\
    \textsc{Human\&Machine}  & 192 $\pm$ 40 & 101 $\pm$ 17\\
    \textsc{Human-only}  & 192 $\pm$ 43 & 100 $\pm$ 16\\
    \textsc{Machine-only} & 199 $\pm$ 58 & 105 $\pm$ 22\\ 
    First drafts & 202 $\pm$ 56 & 104 $\pm$ 22\\
    \bottomrule
    \end{tabular}
 \caption{Statistics of the drafts. The scores are averaged over the drafts. The values following ``$\pm$" denote the standard deviation of the scores. }
 \label{tbl:stats}
\end{table*}

\section{Details on the evaluation in Section~\ref{subsec:exp:quality}}
\label{app:eval}
We used BLEURT-Base with 128 max tokens.\footnote{\url{https://storage.googleapis.com/bleurt-oss/bleurt-base-128.zip}}
BLEURT is designed to evaluate the similarity of a given sentence pair.
Thus, we first split each draft into sentences, and each sentence in first drafts is aligned with the most similar sentence in the corresponding final draft.
Sentence splitting and sentence alignment is achieved by spaCy.\footnote{Sentence similarity is computed using cosine similarity of average word vectors. We used spaCy's \texttt{en\_core\_web\_lg} model.}
Note that the references has been created so that the sentence separation does not change from the original first draft.
Finally, we calculate each sentence pair with BLEURT, and averaged the results.

\end{document}